\documentclass[10pt,twocolumn,letterpaper]{article}

\usepackage{cvpr}              % To produce the CAMERA-READY version
\definecolor{cvprblue}{rgb}{0.21,0.49,0.74}
\usepackage[pagebackref,breaklinks,colorlinks,allcolors=cvprblue]{hyperref}

\def\paperID{29} % *** Enter the Paper ID here
\def\confName{CVPR}
\def\confYear{2025}

\title{Human-centered Interactive Learning via MLLMs for Text-to-Image \\Person Re-identification}

\usepackage[accsupp]{axessibility}
\usepackage{xcolor} 
\usepackage{colortbl} 
\usepackage{algorithm}
\usepackage{algorithmic}
\usepackage{multirow}
\usepackage{array}  
\usepackage{pifont} 
\usepackage{bbding}

\author{
    {Yang Qin$^1$\quad Chao Chen$^2$\quad Zhihang Fu$^2$\quad Dezhong Peng$^{1,4,5}$\quad Xi Peng$^{1,3}$\quad Peng Hu$^1$\thanks{Corresponding author: Peng Hu (penghu.ml@gmail.com).}} \\
    {\normalsize $^1$College of Computer Science, Sichuan University\quad \normalsize$^2$Independent Researcher}\\ {\normalsize$^3$National Key Laboratory of Fundamental Algorithms and Models for Engineering Simulation,  Sichuan University} \\ 
    {\normalsize$^4$Sichuan National Innovation New Vision UHD Video Technology Co., Ltd \quad $^5$Tianfu Jincheng Laboratory} 
}

\begin{document}
\maketitle

\begin{abstract}
Despite remarkable advancements in text-to-image person re-identification (TIReID) facilitated by the breakthrough of cross-modal embedding models, existing methods often struggle to distinguish challenging candidate images due to intrinsic limitations, such as network architecture and data quality. To address these issues, we propose an \textcolor{red}{\textbf{\underline{I}}}nteractive \textcolor{red}{\textbf{\underline{C}}}ross-modal \textcolor{red}{\textbf{\underline{L}}}earning framework (ICL), which leverages human-centered interaction to enhance the discriminability of text queries through external multimodal knowledge. To achieve this, we propose a plug-and-play Test-time Humane-centered Interaction (THI) module, which performs visual question answering focused on human characteristics, facilitating multi-round interactions with a multimodal large language model (MLLM) to align query intent with latent target images. Specifically, THI refines user queries based on the MLLM responses to reduce the gap to the best-matching images, thereby boosting ranking accuracy. Additionally, to address the limitation of low-quality training texts, we introduce a novel Reorganization Data Augmentation (RDA) strategy based on information enrichment and diversity enhancement to enhance query discriminability by enriching, decomposing, and reorganizing person descriptions. Extensive experiments on four TIReID benchmarks, i.e., CUHK-PEDES, ICFG-PEDES, RSTPReid, and UFine6926, demonstrate that our method achieves remarkable performance with substantial improvement. Code is available at \href{https://github.com/QinYang79/ICL}{\textcolor{red}{https://github.com/QinYang79/ICL}}.

\end{abstract}

\begin{figure}[t]
    \centering
    \setlength{\abovecaptionskip}{0.1cm}
    \resizebox{\linewidth}{!}{ 
    \includegraphics{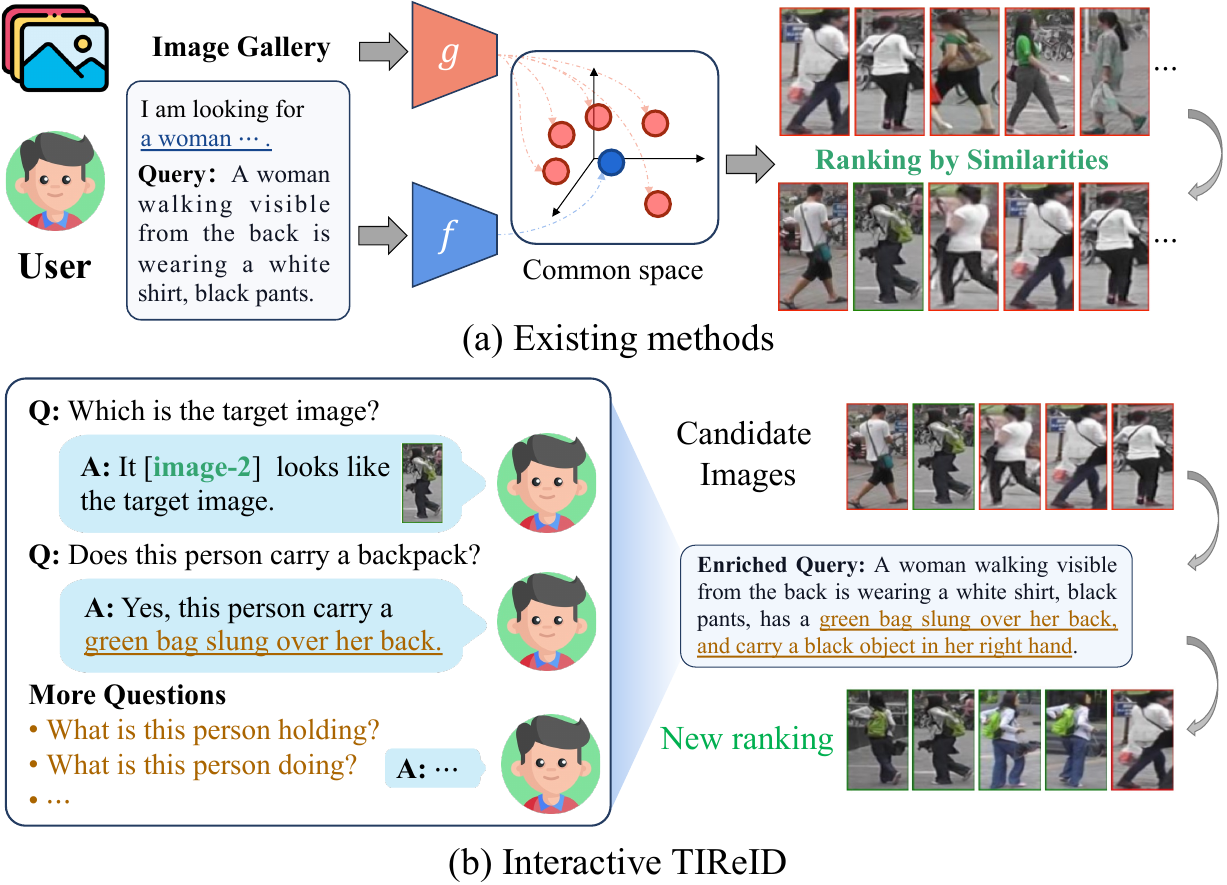}}
    \caption{
   The illustration of our motivation. When performing text-based person re-identification, (a) existing methods commonly exploit cross-modal models to calculate similarity (such as IRRA~\cite{jiang2023cross} and RDE~\cite{qin2024noisy}) and then obtain the candidate person images by ranking. However, due to the intrinsic limitations of models and training data, they may not be able to distinguish challenging candidate images well enough to obtain satisfactory results. (b) Our motivation is to interact with the system for external guidance like a human and gradually refine the user’s query from the candidate items shown in (a) by multiple rounds of question-answering, ultimately improving the overall ranking.}
    \label{fig0}
\vspace{-0.6cm}
\end{figure}

\section{Introduction}
Recently, text-to-image person re-identification (TIReID) \cite{cao2024empirical,ding2021semantically,wang2022caibc,chen2022tipcb,zhu2021dssl,wang2020vitaa,gao2024semi} has made great progress in aligning text descriptions and person images, enabling high-accuracy person search and identification. Different from traditional image-to-image re-identification~\cite{Zhang_2024_CVPR,zuo2023plip,zuo2024cross,wang2025DeMo}, TIReID, as a rising task in the multimodal community~\cite{qin2024dual,qin2025latent,qin2023cross,qin2022deep,sun2024robust}, retrieves person images using user-customized text queries, which is more practical in many scenarios where an image query is unavailable, such as in video surveillance systems~\cite{bukhari2023language} or crowd management~\cite{galiyawala2021person}. A primary challenge in TIReID is learning to associate text queries with the corresponding person images at a fine-grained level, bridging the modality gap for accurate similarity measurement.

To overcome this challenge, recent efforts have developed various strategies to improve cross-modal associations, including local attribute modeling~\cite{zuo2024ufinebench,yan2023clip}, loss function designing~\cite{jiang2023cross,qin2024noisy}, ReID-domain pre-training~\cite{tan2024harnessing,yang2023towards}, \etc. Although these methods achieve promising performance, they are limited by the inherent defect of the offline models and training data, making it hard to handle dynamic user query inputs and leading to poor generalization. In practice, however, different users tend to input concise, vague, and diverse text queries based on their memories to the target person, thereby raising challenges for the offline models to understand fine-grained human characteristics from the queries. For example, as illustrated in~\Cref{fig0} (a), the user text query  ``\emph{A woman walking visible from the back is wearing a white shirt, and black pants.}'' fails to capture some important contextual details like \underline{handheld objects} or \underline{background}, resulting in inaccurate candidate images. In brief, it is hard for offline models to handle dynamic and challenging queries solely relying on their learned internal knowledge. To break through this bottleneck, we could seek to align queries with image candidates with the help of external knowledge, thereby reranking the retrieved images and boosting the identification accuracy.

Recent studies~\cite{liimage,levy2024chatting,han2024merlin} have explored integrating external knowledge from pre-trained models or (multimodal) large language models to guide various downstream tasks. For instance, Li \etal.~\cite{liimage} leveraged multimodal knowledge of pre-trained vision-language models~\cite{radford2021learning} to enhance unimodal image representations for external-guided clustering. However, this approach cannot achieve post-hoc improvement for offline models, leading to infeasibility in practice due to the prohibitive cost of fine-tuning or even retraining. In cross-modal retrieval, Han~\etal.~\cite{han2024merlin} exploited the excellent comprehension and generation capabilities of multimodal large language models (MLLMs) to design an interactive re-ranking pipeline for text-to-video retrieval. However, Han \etal.~\cite{han2024merlin} introduces label leakage by using the ground truth video as the video in mind for the MLLM answerer agent, which is not applicable in actual testing.

In this paper, we present a novel Interactive Cross-modal Learning framework (ICL) that exploits the multimodal knowledge implicit in MLLMs to enhance the alignment between text queries and target person images. Specifically, we first propose a plug-and-play Test-time Human-centered Interaction (THI) module that refines text queries through interactions with an MLLM, enhancing the post-hoc ability of trained models to distinguish challenging candidates. THI identifies the latent target images through multi-round interactions with MLLMs, asking human-centered questions to the MLLM for fine-grained answers about the images, which refine the query texts to strengthen the alignment with the images, ultimately address the inherent limitations of the input queries and enhancing ranking accuracy. To address the intrinsic limitations in the trained models, we propose a Reorganization Data Augmentation strategy (RDA) to enrich and diversify pedestrian texts for training enhancement, thereby transferring external knowledge in MLLMs into the model. To enrich the texts, RDA applies visual question answering via the MLLM on each image-text pair to supplement fine-grained person characteristics. Moreover, to enhance the diversity of texts, RDA presents a decomposition-reorganization strategy to decompose person descriptions into attribute-specific sub-sentences (\eg, clothes, pants, shoes, \etc.), and rewrite them with MLLM into multiple sentences with the same meaning. Subsequently, the rewritten sentences are randomly reordered and recombined to generate varied augmented texts, thus boosting data diversity and enhancing the generalization of models. Our main contributions are as follows:
\begin{itemize}
    \item We introduce human-centered interaction to TIReID, proposing a novel MLLM-driven Interactive Cross-modal Learning framework (ICL), which leverages external knowledge to overcome the inherent limitations of existing offline methods in handling dynamic queries.
    \item A plug-and-play Test-time Human-centered Interaction module (THI) is presented to align query intent with latent target images through multi-round interactions with MLLMs, improving the ranking quality.    
    \item An effective Reorganization Data Augmentation strategy (RDA), applying MLLMs for text decomposition and recombination, is developed to generate discriminative and diverse training texts, improving cross-modal learning.
    \item Extensive experiments on four text-to-image person re-identification benchmarks verify the effectiveness and superiority of our method, which achieves promising performance and superior generalization.
\end{itemize}

\begin{figure*}[t]
    \centering
    \setlength{\abovecaptionskip}{0.1cm}
    \resizebox{0.95\linewidth}{!}{ 
    \includegraphics{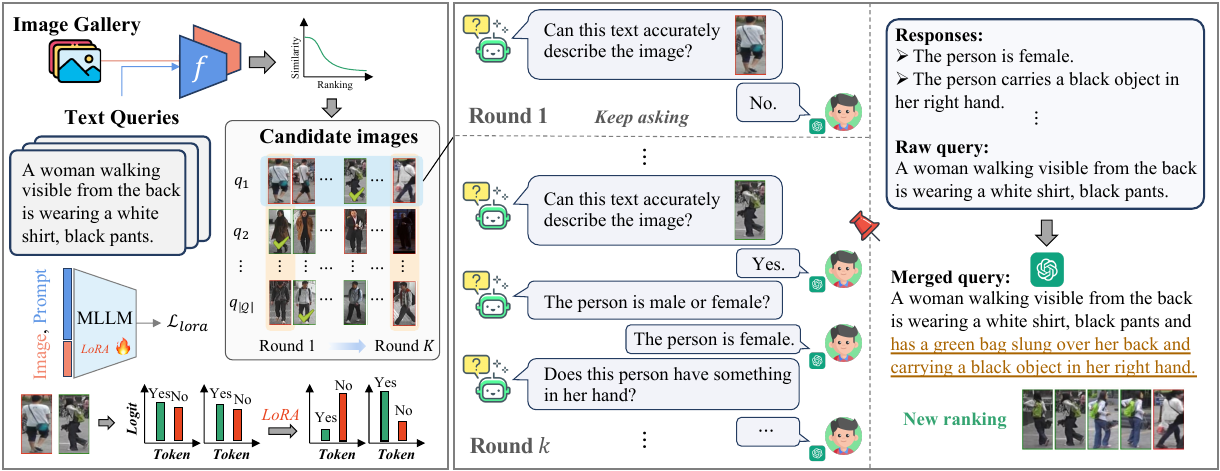}}
    \caption{ The illustration of our Test-time Human-centered Interaction (THI) module. THI includes $K$ rounds of interactions to align query intention with the latent target image by external guidance, where in each round, we perform human-centered visual question answering around fine-grained person attributes to enhance the semantic consistency between the query and the intended person image, and then improve the final ReID performance on the large-scale evaluation through efficient re-ranking. Besides, we perform supervised fine-tuning via LoRA to inspire the discriminate ability of MLLM for ReID domain images and better align queries with latent target images.}
    \label{fig_tui} 
    \vspace{-0.3cm}
\end{figure*}

\section{Related Work}
\subsection{Text-to-Image Person Re-identification}
As a challenging topic in multimodal learning, text-to-image person re-identification (TIReID)  aims to search the image of the target person with a given natural language query. Existing methods~\cite{zhu2021dssl,bai2023rasa,jiang2023cross,qin2024noisy,tan2024harnessing} can be roughly divided into three categories according to the used backbone type: unimodal backbones, general multimodal pre-trained backbones, and ReID-domain multimodal pre-trained backbones. The early TIReID methods~\cite{wang2020vitaa,wu2021lapscore} commonly use modality-specific unimodal backbones to encode text or images for cross-modal alignment, \eg, ResNet~\cite{he2016deep}, BERT~\cite{wang2024spectrum}, \etc.  Recently, benefiting from the rapid development of visual-language pre-training models~\cite{radford2021learning,li2022blip}, more and more researchers~\cite{jiang2023cross,cao2024empirical,qin2024noisy} have begun to use general pre-trained models as the backbone networks for solving TIReID, hoping that the pre-trained alignment knowledge can improve modality representation and cross-modal alignment. However, the performance improvement is still limited due to the domain gap between the general pre-training data and TIReID tasks. To this end, some attempts~\cite{yang2023towards,tan2024harnessing} exploit image-caption models or MLLMs to annotate images, thus obtaining a large number of image-text pairs for ReID-domain pre-training. Although promising performance has been achieved, they still cannot get rid of the inherent defect of offline models during handling dynamic queries. 
In this paper, we propose a novel interactive cross-modal learning framework based on MLLMs, which exploits dynamic interactions for external guidance to improve the generalization ability.

\subsection{Visual Interactive Learning}
With the booming development of multimodal large language models (MLLMs)~\cite{liu2023llava,li2023blip,Qwen2VL}, exploiting visual interaction to improve downstream tasks has attracted increasing attention from researchers. We collectively refer to these as visual interaction learning and divide them into offline interaction~\cite{wang2024multimodal,tan2024harnessing} and online interaction~\cite{levy2024chatting,han2024merlin,he2025chatting} groups by engaging in specific task periods. The former aims to use interactions to improve the training/pre-training data, such as information richness and diversity. It is usually static and separated from users, which is the mainstream of existing TIReID methods, \eg, Tan \etal.~\cite{tan2024harnessing} apply MLLMs to conduct interactions with multiple prompt templates for better diversity. Unlike them, online interactions are designed to be able to use real-time feedback to improve training or testing.  For example, Levy \etal.~\cite{levy2024chatting} propose to gradually clarify user intention through dialogue to improve image retrieval performance. Likewise, Han \etal.~\cite{han2024merlin} develop a rerank pipeline based on LLM-based iterative navigation. However, these methods cannot be directly used for TIReID effectively due to the differences in tasks and data domains. In this paper, we develop a test-time module based on MLLMs to conduct interactions for external guidance, overcoming the limitations of the intra-model knowledge in TIReID offline models.

% Multimodal human-machine interactions begin with visual question answering (VQA) and related dialogue tasks, which aim to obtain additional clear information by leveraging interactions, thus improving specific tasks, \eg, visual Conversations and visual Search, \etc.

\section{Methodology}
In this section, we introduce an Interactive Cross-modal Learning framework (ICL), which consists of two core comments to address the inherent challenges in offline models and training data, \ie, Test-time Human-centered Interaction module (THI) and Reorganization Data Augmentation (RDA). In~\Cref{sec3.1}, we provide the necessary definitions to facilitate the study. Then, we outline the details of our THI and RDA in~\Cref{sec3.2,sec3.3}, respectively. 

\subsection{Problem and Symbol Statement \label{sec3.1}}

Suppose that we have the query text $q\in\mathcal{Q}$, a pedestrian image $v\in\mathcal{V}$, where $\mathcal{Q}$ and $\mathcal{V}$ are the text query set and image gallery. The purpose of TIReID is to utilize the text query to retrieve the ideal images from the gallery, thus achieving person searching by identities of retrieved images. To achieve this, existing methods usually train an offline cross-modal model $f_\text{cross}=\{f,g\}$ to measure the similarities between text queries and pedestrian images. For a text $q$ and an image $v$, the similarity to measure the matching degree can be represented as
$S_{q,v}\equiv \operatorname{Sim}(f(q), g(v))$,
where $\operatorname{Sim}(*)$ is the similarity function, $f$ and $g$ are the text and image encoders, respectively.
% obtain cross-modal sample representations, \ie,
% \begin{equation}
%     \mathbf{e}_q = f(q),\quad \mathbf{e}_v = g(v),
% \end{equation}
% where $\mathbf{e}_q$ and $\mathbf{e}_v$ are the embedding vectors in IRRA~\cite{jiang2023cross} or multi-granularity embedding vectors like in RDE~\cite{qin2024noisy}. Then, calculate the similarity to measure the matching degree between text and image, \ie, $S_{q,v}\equiv \operatorname{Sim}(\mathbf{e}_q,\mathbf{e}_v)$, where $\operatorname{Sim}(*)$ is the similarity function (\eg, cosine
% similarity, \etc.). 
Then, we can utilize $q$ to search relevant candidate images $\hat{v}$ from $\mathcal{V}$ as follows:
\begin{equation}
   \hat{\mathcal{V}}(q)  = \{\hat{v}_k\}^K_{k=1} = \text{Top-}K_{v\in\mathcal{V}} (S_{q,v}),
   \label{eq2}
\end{equation}
 where $\hat{\mathcal{V}}(q)$ is the candidate set for $q$ and $K$ is the number of candidate images. In addition, our ICL also involves MLLMs to conduct interactions and training augmentation, which we denote as $\mathcal{M}$, and the prompt template function for $\mathcal{M}$ is denoted by the symbol $\mathcal{T}$. Based on the above definitions, we will elaborate on ICL in the following sections.
    
\subsection{Test-time Humane-center Interaction \label{sec3.2}} 
\textbf{Anchor Localization.} Due to the model bottleneck, it is hard for existing methods to distinguish challenging candidate images relying solely on internal knowledge. Thus, we exploit the fine-grained image understanding capability of MLLMs to conduct external guidance. We first propose asking questions such as “\emph{Can this text accurately describe the image?}” to let MLLMs explicitly tell us the latent target (\underline{anchor}) image. As shown in~\Cref{fig_tui}, given a text query $q$, we first get the corresponding candidate set $\hat{\mathcal{V}}(q)$ and then perform multiple ($K$) rounds of interactions according to the ranking until the ideal image is determined. For the $k$-th round, the answer for interaction can represented as:
\begin{equation}
    a^q_{\hat{v}_k}  = \mathcal{M}(\mathcal{T}_\text{loc} (q,\hat{v}_k)),
    \label{eq_al}
\end{equation}
where $k\in\{1,\cdots,K\}$, $a^q_{\hat{v}_k}$ is the answer of `Yes' or `No', $\hat{v}_k$ is the Top-$k$ candidate image for query $q$, and $\mathcal{T}_\text{loc}$ is the prompt template function for anchor localization. 
However, since there is a domain gap between ReID images and the generic images used for pre-training or instruction-tuning MLLMs, the answers of MLLMs are often unreliable. To solve this, we exploit LoRA~\cite{hu2021lora} to perform supervised fine-tuning (SFT) on MLLMs to inspire the fine-grained ability to identify the person image and the SFT loss is:
\begin{equation}
    \mathcal{L}_{lora}= - \sum_{(x,y)\in\mathcal{Z}}\sum^{ |y| }_{t=1} \log (p_{\mathcal{M}}(y_t|x,y<t)),
\end{equation}
where $x$ is the input \texttt{Pormpt} and $y$ is the output \texttt{Response}, $\mathcal{Z}=\left\{\mathcal{Z}^+,\mathcal{Z}^-\right\}$ is the SFT dataset. For $\mathcal{Z}^+$, we select part of training texts ($\{q_i\}^{N_l}_{i=1}$) and the corresponding ground-truth images  ($\{v^+_i\}^{N_l}_{i=1}$) from the training set to construct the input \texttt{Pormpts}, \ie, $\{ \mathcal{T}_\text{loc} (q_i,v^+_i)\}^{N_l}_{i=1}$, expecting MLLMs to output the \texttt{Response} of `Yes'.
Similarly, for $\mathcal{Z}^-$, we also use the text queries and the corresponding negative images from the training set to construct the input \texttt{Pormpts}, expecting MLLMs to output the \texttt{Response} of `No'. To make $\mathcal{Z}^-$ more discriminative, we use a TIReID pre-trained cross-modal model to obtain the Top-$10$ image with different person ID as input negative sample image for each text query by similarity ranking.

\noindent\textbf{Human-centered VQA.} For each round, once we confirm the response of `Yes', we will ask the anchor image ($\bar{v}$) a series of questions to learn more details about pedestrian characteristics. These details can help us alleviate the discrepancy between the user query and the information within the anchor image, thus improving overall ranking by refining the query. We call this process human-centered visual question answering (VQA), which is expressed as:
\begin{equation}
    r_{\bar{v}}=\mathcal{M}(\mathcal{T}_\text{vqa} (\{c_i\}^{N_q}_{i=1},\bar{v})),
    \label{eq5}
\end{equation}
where $\{c_i\}^{N_q}_{i=1}$ are $N_q$ questions directed at the detail characteristics (\eg, gender, hair, upper body, lower body, shoes, \etc.) of the pedestrian image, $r_{\bar{v}}$ is the answer to $\{c_i\}^{N_q}_{i=1}$  one by one, and $\mathcal{T}_\text{vqa}$ is the prompt template function for human-centered VQA.
To improve the consistency between the image and the user query, a natural strategy is to concatenate these one-by-one answers after the original text. However, due to the limitation of the maximum text length that can be processed by the text model (\eg,  it is usually $77$ for CLIP), this will destroy the sequence structure. To this end, we recommend merging these answers and the original text by MLLMs to obtain a fluent and concise text, \ie,
\begin{equation}
    \hat{q}=\mathcal{M}(\mathcal{T}_\text{aggr} (r_{\bar{v}},q)),
    \label{eq6}
\end{equation}
where $\hat{q}$ is the merged text query and $\mathcal{T}_\text{aggr}$ is is the prompt template function for text aggregation.

\begin{figure}[t]
\centering
\setlength{\abovecaptionskip}{0.1cm}
\begin{subfigure}{0.325\linewidth} 
\includegraphics[height=0.73\linewidth]{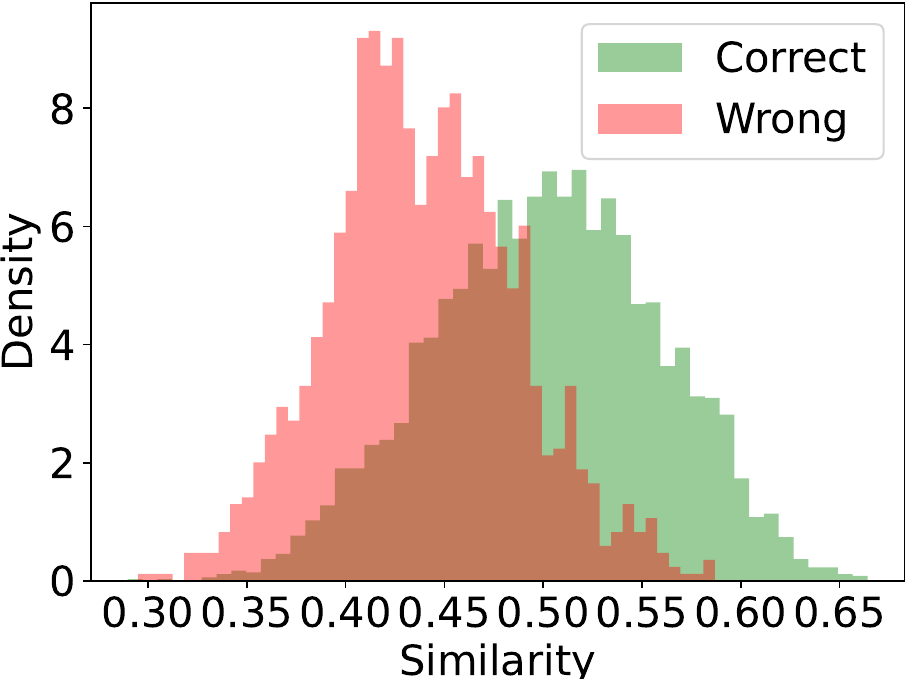}
\centering
\caption{\footnotesize {CUHK-PEDES}}
\label{cuhk_test_sims}
\end{subfigure}
\begin{subfigure}{0.325\linewidth} 
\includegraphics[height=0.73\linewidth]{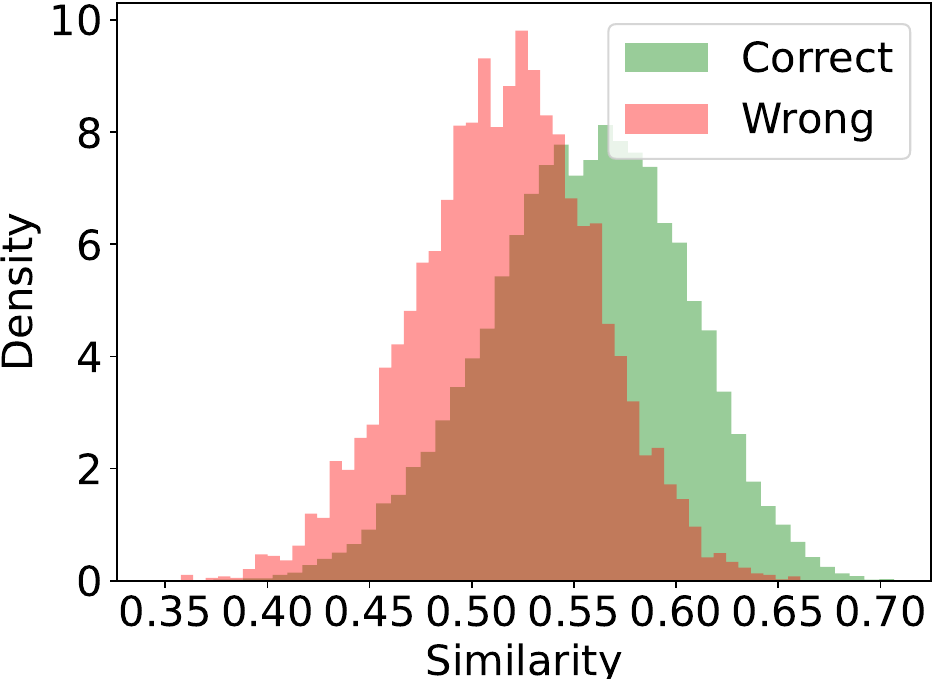}
\centering
\caption{\footnotesize  {ICFG-PEDES}}
\label{icfg_test_sims}
\end{subfigure}
\begin{subfigure}{0.325\linewidth} 
\includegraphics[height=0.73\linewidth]{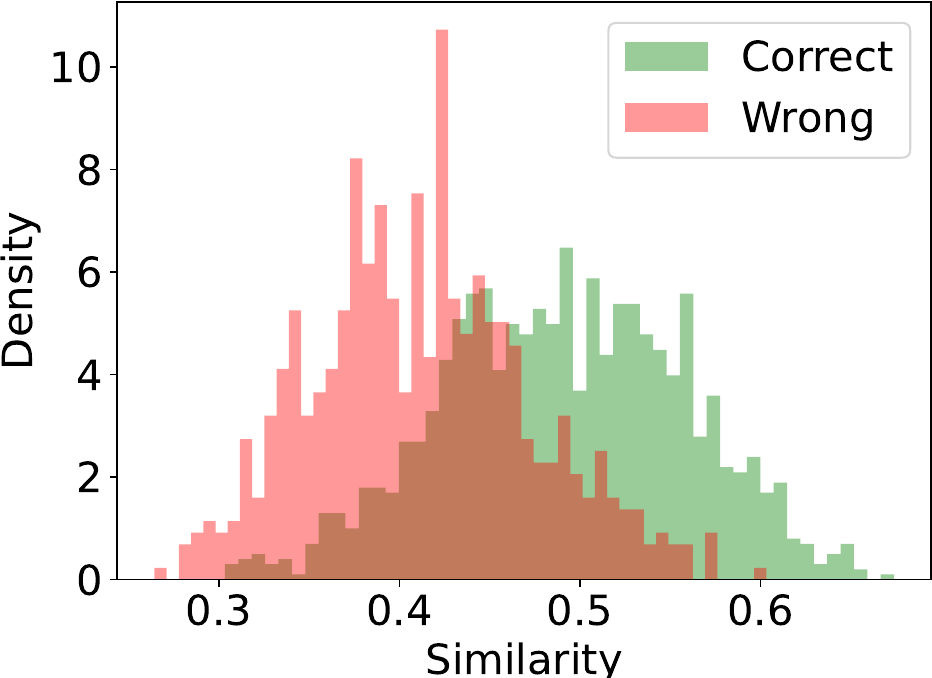}
\centering
\caption{\footnotesize  {RSTPReid}}
\label{rstp_test_sims}
\end{subfigure}
\caption{The similarity statistics of the Top-$1$ retrieved items in the test sets of the three benchmarks. It is obvious that the items with higher similarity are more likely to be correct retrievals.} %还没替换
\label{fig_sims} 
\vspace{-0.3cm}
\end{figure}

\noindent\textbf{Efficient Re-ranking.} To apply the above process to large-scale evaluation, efficiency is a factor that needs to be considered in the retrieval task, since introducing MLLMs to interactions for external guidance inevitably brings additional inference costs. To improve the interaction efficiency, we recommended adopting different strategies for different rounds of interactions. We observe that the retrieval quality is positively correlated with the similarity between the query and the Top-$1$ candidate image, as shown in~\Cref{fig_sims}. 
To this end, in the first round, we only interact with images whose cross-modal similarity is greater than a threshold $\xi$ and the answer of anchor localization is `Yes'. This is because the retrievals with low similarities are more likely to be wrong items. After the first round, we only interact with the image that is all considered `No' in the previous rounds of anchor localization and whose cross-modal similarity is less than the threshold $\xi$. This allows us to reduce unnecessary interactions and find text queries that really require interactions, making our method more efficient on large-scale evaluation. Given a query $q$, if the above constraints are met, the interaction is carried out and we can get merged text $\hat{q}$ after human-centered VQA, the re-ranking similarity $\hat{S}_{q,v}$ of any image $v$ in $\mathcal{V}$ is:
\begin{equation}
   \hat{S}_{q,v} =  \lambda S_{q,v} + (1-\lambda)\bar{S}_{\hat{q},v},
   \label{eq7}
\end{equation} 
where $\bar{S}_{\hat{q},v} \equiv 1$ if $v$ is $\hat{v}_1$, otherwise, $\bar{S}_{\hat{q},v}\equiv{S}_{\hat{q},v}$, and $\lambda\in[0,1]$ is a hyperparameter to balance the contribution of the raw query and the refined text. To make our approach clearer, we describe the detailed algorithm process of our THI in~\Cref{Impl}. Due to space limitations, all prompt templates can be found in the supplementary material.

\begin{algorithm}[!htbp]
	\renewcommand{\algorithmicrequire}{\textbf{Input:}}
	\renewcommand{\algorithmicensure}{\textbf{Output:}}
	\caption{The interaction process of our THI}
	\label{Alg1}
	\begin{algorithmic}[1]
	\REQUIRE The query set $\mathcal{Q}$, the image gallery $\mathcal{V}$, the offline model $f_\text{cross}$, the MLLM $\mathcal{M}$, the similarity threshold $\xi$, the number of interaction rounds $K$;  
    \STATE Obtain candidate sets $\{\hat{\mathcal{V}}(q_i) \}^{|\mathcal{Q}|}_{i=1}$ for all queries in $\mathcal{Q}$ via~\Cref{eq2};
    \FOR{ $k = 1, 2,\cdots,K$}
        \FOR{ $i = 1, 2,\cdots,|\mathcal{Q}|$}
           \STATE Conduct anchor localization via~\Cref{eq_al} and output the answer of $a^{q_i}_{\hat{v}_{k}^i}$ based on the $k$-th candidate image $\hat{v}_{k}^i$ in $\hat{\mathcal{V}}(q_i)$;
           \IF{$a^{q_i}_{\hat{v}_{k}^i}$ shows `Yes', $k=1$, and $S_{q_i,\hat{v}_1^i}>\xi$}
           \STATE Conduct human-centered VQA via \Cref{eq5,eq6} to get the refined query $\hat{q}_i$;
           \STATE Compute the re-ranking similarities between query $q_i$ and all images via \Cref{eq7};
           \ENDIF
          \IF{$\{a^{q_i}_{\hat{v}_{j}^i}\}^{k-1}_{j=1}$ all show `No', $a^{q_i}_{\hat{v}_{k}^i}$ shows `Yes', $k>1$, and $S_{q_i,\hat{v}_1^i}\leq\xi$}
           \STATE Conduct human-centered VQA via \Cref{eq5,eq6} to get the refined query $\hat{q}_i$;
           \STATE Compute the re-ranking similarities between query $q_i$ and all images via \Cref{eq7};
           \ENDIF 
        \ENDFOR
    \ENDFOR 
     \STATE Re-ranking based on similarities;
    \ENSURE The new candidate images.
	\end{algorithmic}  
\label{Impl}
\end{algorithm}

\subsection{Reorganization Data Augmentation \label{sec3.3}}

\begin{figure}[t]
    \centering
    \setlength{\abovecaptionskip}{0.1cm}
    \resizebox{\linewidth}{!}{ 
    \includegraphics{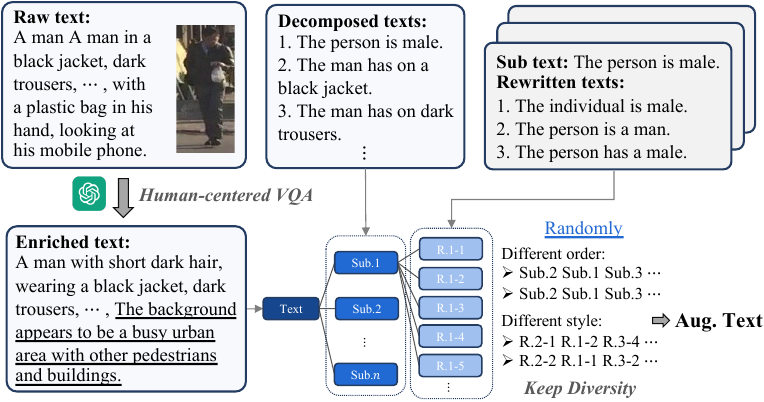}}
    \caption{
    The illustration of our RDA. The purpose of RDA is to supplement more details to the original training texts through human-centered VQA, improving the discriminability of texts. In addition, to enhance diversity, RDA maximizes diversity through the Decomposition-Rewriting-Reorganization strategy.}
    \label{fig_idn}
    % \vspace{-0.6cm}
\end{figure}

Although THI can exploit interactions to provide external knowledge during test time to improve retrieval quality, the cross-modal embedding model is still a bottleneck that limits its further performance improvement. To this end, as shown in~\Cref{fig_idn}, we introduce a new Reorganization Data Augmentation (RDA) strategy by enriching, decomposing, and reorganizing person descriptions to improve the discriminability and diversity of training data. We first apply Human-centered VQA in THI to obtain relevant characteristic descriptions and merge them with the original text, thus obtaining a text with richer information, \ie, 
$
    \hat{q}=\mathcal{M}(\mathcal{T}_\text{aggr} (\mathcal{M}(\mathcal{T}_\text{vqa} (\{c_i\}^{N_q}_{i=1},v)),q))
$. Then, we apply MLLMs to decompose enriched text $\hat{q}$ into multiple independent sub-sentences that describe individual attributes without interfering with each other, \ie, $\{ \tilde{q}_i \}^{n}_{i=1} = \mathcal{M}(\mathcal{T}_\text{dec}(\hat{q}))$, where $\{ \tilde{q}_i \}^{n}_{i=1}$ are the $n$ sub-sentences and $\mathcal{T}_\text{dec}$ is the prompt template function for text decomposition. The purpose of decomposition is to be able to reorganize the text in a different order later. To increase diversity, we rewrite each sub-sentence into multiple sentences with different styles but the same meaning. For sub-sentence $\tilde{q}$, we conduct rewriting by $\mathcal{R} = \{ \breve{q}_j \}^{m}_{j=1} = \mathcal{M}(\mathcal{T}_\text{rwt}(\tilde{q}))$, where $\mathcal{R}$ is the set with $m$ rewritten sentences and $\mathcal{T}_\text{rwt}$ is the prompt template function for text rewriting.

Finally, for a text $q$, we can obtain the set $\{\mathcal{R}_i\}^n_{i=1}$ that contains a large number of style sub-sentences. So far, we can obtain augmented texts with different sub-sentence orders and different style combinations as shown in~\Cref{fig_idn}, which we represent as $\check{q}$. During training, we can mix the augmentation texts ($\check{q}$) with the original texts ($q$) for cross-modal learning, thereby improving model-intra knowledge and generalization. Due to space limitations, more training details can be found in the supplementary material.     

\begin{table*}[t]
\centering 
\setlength{\abovecaptionskip}{0.1cm}
\arrayrulecolor{black} 
\resizebox{\textwidth}{!}{
\setlength\tabcolsep{1.5pt}
\begin{tabular}{l|cc|ccccc|ccccc|ccccc}\hline\hline
\rowcolor{gray!25}   &    &    & \multicolumn{5}{c|}{{CUHK-PEDES}}       & \multicolumn{5}{c|}{{ICFG-PEDES}}       & \multicolumn{5}{c}{{RSTPReid}}                          \\
 \rowcolor{gray!25} Methods &Image Enc.&Text Enc.  & Rank-1            & Rank-5           & Rank-10  & mAP   & mINP  & Rank-1   & Rank-5   & Rank-10  & mAP   & mINP & Rank-1         & Rank-5        & Rank-10       & mAP   & mINP  \\\hline
\multicolumn{18}{l}{\textcolor{cyan}{\ding{182}} \emph{VL-Backbones w/o ReID-domain pre-training}} \\\hline
IVT~\cite{shu2022see}      & ViT-Base   & BERT         & 65.69          & 85.93         & 91.15 & 60.66 & -     & 56.04 & 73.60 & 80.22 & -     & -    & 46.70       & 70.00      & 78.80      & -     & -     \\
LCR$^2$S~\cite{yan2023learning}    & RN50       & BERT         & 67.36          & 84.19         & 89.62 & 59.20 & -     & 57.93 & 76.08 & 82.40 & 38.21 & -    & 54.95       & 76.65      & 84.70      & 40.92 & -     \\                                       
% UniPT~\cite{shao2023unified}    & ViT-Base   & BERT         & 68.50          & 84.67         & 90.38 & -     & -     & 60.09 & 76.19 & 82.46 & -     & -    & 51.85       & 74.85      & 82.85      & -     & -     \\
CFine~\cite{yan2023clip}    & CLIP-ViT   & BERT         & 69.57          & 85.93         & 91.15 & -     & -     & 60.83 & 76.55 & 82.42 & -     & -    & 50.55& 72.50& 81.60 & -     & -     \\
RaSa~\cite{bai2023rasa}    & Swin-B     & BERT         & 76.51          & 90.29         & 94.25 & 69.38 & -     & 65.28 & 80.40 & 85.12 & 41.29 & -    & 66.90       & 86.50      & 91.35      & 52.31 & -     \\ 
% MARS~\cite{ergasti2024mars}     & Swin-B     & BERT         & 77.62          & 90.63         & 94.27 & 71.71 & -     & 67.60 & 81.47 & 85.79 & 44.93 & -    & 67.60       & 81.47      & 85.79      & 44.93 & -     \\

IRRA~\cite{jiang2023cross}     & CLIP-ViT   & CLIP-X. & 73.38          & 89.93         & 93.71 & 66.13 & 50.24 & 63.46 & 80.25 & 85.82 & 38.06 & 7.93 & 60.20       & 81.30      & 88.20      & 47.17 & 25.28 \\
TBPS~\cite{cao2024empirical}     & CLIP-ViT   & CLIP-X. & 73.54          & 88.19         & 92.35 & 65.38 & 49.25     & 65.05 & 80.34 & 85.47 & 39.83 & 7.87    & 62.10& 81.90& 87.75& 48.00& 25.86
% 61.95       & 83.55      & 88.75      & 48.26 & 25.86   

\\
CFAM~\cite{zuo2024ufinebench}	&CLIP-ViT  	&CLIP-X.	&75.60 &	{90.53}& 	94.36 &	67.27 &	-	&65.38&	81.17 &	86.35 	&39.42 &	-	&62.45& 	83.55 	&91.10 	&49.50& 	-	\\
RDE~\cite{qin2024noisy}  & CLIP-ViT   & CLIP-X. & 75.94          & 90.14         & 94.12 & 67.56 & 51.44 & 67.68 & 82.47 & 87.36 & 40.06 & 7.87 & 65.35       & 83.95      & 89.90      & 50.88 & 28.08 \\\hline
% \rowcolor{green!10} \textbf{Our ICL}& CLIP-ViT   & CLIP-X. &76.15&90.42&{94.43}&67.78&51.74&67.95&82.68&87.41&41.11&8.43&67.75&{85.85}&{91.65}&52.61&29.03 \\
% \rowcolor{red!10}\textbf{Our ICL$\star$}& CLIP-ViT   & CLIP-X. & {77.11}&90.20&94.28&{68.69}&{53.14} &69.20&82.48&87.26&41.48&8.57&    {70.40}&85.50&91.55&{53.77}&{30.12}\\
% \hline
\rowcolor{green!10} \textbf{Our ICL}& CLIP-ViT   & CLIP-X. &76.41&90.48&{94.33}&68.04&51.99&68.11&82.59&87.52&40.81&8.18&67.70&{86.05}&{91.75}&52.62&29.36 \\
\rowcolor{red!10}\textbf{Our ICL$\star$}& CLIP-ViT   & CLIP-X. &77.91&90.27&94.14&69.13&53.40&69.02&82.45&87.36&41.21&8.30&70.55&85.95&91.65&53.68&30.13\\
\hline

\multicolumn{18}{l}{\textcolor{cyan}{\ding{183}} \emph{VL-Backbones with ReID-domain pre-training}} \\\hline
IRRA$^\flat$~\cite{jiang2023cross} & CLIP-ViT   & CLIP-X. &74.05&89.48&93.64&66.57&-&64.37&80.75&86.12&38.85&-&61.90 &	80.60&89.30&48.08&-\\
APTM~\cite{yang2023towards}     & Swin-B     & BERT         & 76.53          & 90.04         & 94.15 & 66.91 & -    & 68.51       & 82.99      & 87.56      & 41.22 & -   & 67.50 & 85.70 & 91.45 & 52.56 & -       \\
% NAM~\cite{tan2024harnessing} & CLIP-ViT   & CLIP-X. & 76.82          & 91.16         & 94.46 & 69.55 & -     & 67.05 & 82.16 & 87.33 & 41.51 & -    & 68.50       & \textbf{87.15}      & \textbf{92.10}      & 53.02 & -     \\
NAM$^\natural$~\cite{tan2024harnessing} & CLIP-ViT   & CLIP-X. & 77.47          & 90.84         & 94.67 & 69.43 & 54.08     & 66.76 & 82.02 & 87.17 & 41.45 & 9.53    & 67.15       & {86.55}      & {91.90}      & 52.00 & 28.46     \\
% RDE$^\diamondsuit$~\cite{qin2024noisy} & CLIP-ViT   & CLIP-X. &77.73&91.05&94.80&69.47&53.46&68.83&83.36&88.17&41.48&8.53& 68.85&84.95&90.10&52.59&28.36    \\   
\hline
% NAM+APTM & Swin-B     & BERT         & 78.13          & 91.19         & 94.50 & 68.75 & -     & 69.37 & 83.55 & 88.18 & 42.42 & -    & 69.95       & 87.35      & 92.30      & 54.17 & -   \\
%     \rowcolor{green!10}Our ICL& CLIP-ViT   & CLIP-X. &77.92&91.81& 95.22& 69.68& 53.63& 69.24&83.21&88.01&42.03&8.92 &70.90&85.65&91.05&53.30&28.57 \\
% \rowcolor{red!10}Our ICL$\star$& CLIP-ViT   & CLIP-X. & &&&&&&&&&&72.80&86.45&91.55&54.14&29.31
% \\
%     \rowcolor{green!10}\textbf{Our ICL}& CLIP-ViT   & CLIP-X. &78.22&91.16&94.66&69.36&53.17&69.18&82.98&87.67&42.30&9.09&70.05&87.15&92.15&53.60&29.79\\
% \rowcolor{red!10}\textbf{Our ICL$\star$}& CLIP-ViT   & CLIP-X. &78.70&90.97&94.56&70.09&54.28& 69.82&82.94&87.56&42.66&9.23&71.80&87.10&92.05&54.61&30.56
    \rowcolor{green!10}\textbf{Our ICL}& CLIP-ViT   & CLIP-X. &78.18&91.63&94.83&69.58&53.48&69.22&83.49&88.06&42.34&9.01&70.00&86.60&91.70&54.16&30.93\\
\rowcolor{red!10}\textbf{Our ICL$\star$}& CLIP-ViT   & CLIP-X. &79.06&91.26&94.72&70.44&54.70 &  70.05&83.35&87.91&42.70&9.13 &72.55&86.60&91.30&55.19&31.72

\\
\hline\hline
\end{tabular} 
} 
\caption{Performance on the three coarse-grained benchmarks. The results with THI are marked with $\star$. Note that IRRA$^\flat$ means using the pre-trained Backbones with MALS~\cite{yang2023towards} and the results of NAM$^\natural$ are reproduced by us.} 
\label{tab_main}
\vspace{-0.3cm}
\end{table*}

\section{Experiments}

In this section, we conduct extensive experiments to verify the effectiveness, superiority, and generalization of the proposed ICL on four public benchmark datasets. 
% Our main questions to be explored and verified are as follows:
% \begin{itemize}
%     \item \textbf{Superiority}: Does our ICL have advantages in standard in-domain evaluation? (\Cref{sec4.1})
%     \item  \textbf{Generalizability}: Does our ICL framework improve the cross-domain generalization of the model? (\Cref{sec4.2})
%     \item  \textbf{Transferability}: Whether our THI can be applied to existing methods to improve performance? (\Cref{sec4.3})
%     \item \textbf{Ablation study}: What are the key factors for the performance improvement of our method? (\Cref{sec4.4})
%     \item \textbf{Parameter Analysis and Visualization}: Explore the impact of parameters on our method and intuitively observe the effect of interaction. (\Cref{sec4.5})
% \end{itemize}
\subsection{Datasets and  Evaluation Protocols}
\textbf{Datasets.} In our experiments, we use three coarse-grained benchmarks, CHUK-PEDES~\cite{li2017person}, ICFG-PEDES~\cite{ding2021semantically}, and RSTPReid~\cite{zhu2021dssl}, and one fine-grained benchmark, UFine6926~\cite{zuo2024ufinebench}, to evaluate our ICL. Compared with the coarse-grained benchmark, the fine-grained benchmark has richer texts with fine-grained information. For all datasets, we follow their official settings for data partitioning. More details are provided in the supplementary materials.\\
\textbf{Evaluation Protocols.} To measure performance, we utilize the widely accepted Rank-K (1,5,10) metrics to measure the TIReID performance. Like~\cite{jiang2023cross,qin2024noisy}, we also report the mean Average Precision (mAP) and mean Inverse Negative Penalty (mINP) as auxiliary metrics.

\subsection{Implementation Details}
To achieve interactive TIReID, we choose RDE~\cite{qin2024noisy} and Qwen2-VL-7B-Instruct~\cite{Qwen2VL} as the investigated TIReID method and MLLM. For the sake of fairness, we do not modify any settings of RDE, including model architecture (CLIP-ViTB/16~\cite{radford2021learning}), hyperparameters, and training parameters. To be compatible with the fine-grained benchmark, following~\cite{zuo2024ufinebench}, the maximum length of the textual tokens of CLIP is set to $168$ by interpolating the positional embedding layer with an initial learning rate of 5$e$-5. As for fine-tuning MLLMs, we use the Llama-Factory
framework~\cite{zheng2024llamafactory} to conduct SFT with LoRA~\cite{hu2021lora} for $2$ epochs. The train batch size per device is set to $4$, the gradient accumulation steps are set to $16$, the initial learning rate is 5$e$-5, and the hyperparameters $\alpha$ and $r$ of LoRA are set to $16$ and $8$, respectively. During inference, the temperature is set to $0.01$ to keep reproducibility. Note that all experiments can be completed on two GeForce RTX 24GB 3090 GPUs.

\begin{table}[t]
\centering
\setlength{\abovecaptionskip}{0.1cm}
\resizebox{\linewidth}{!}{
\setlength\tabcolsep{6pt}
\begin{tabular}{l|ccccc}\hline 
 \rowcolor{gray!25} Methods&  Rank-1 &  Rank-5&Rank-10 & mAP &mINP \\\hline  
LGUR~\cite{shao2022learning} & 70.69&  84.57 & 89.91 & 68.93 & - \\ 
SSAN~\cite{ding2021semantically} &75.09& 88.63& 92.84& 73.14 &- \\
IRRA~\cite{jiang2023cross} &85.02&94.31&96.75&83.91&77.30\\
% RDE~\cite{qin2024noisy} &88.49&95.81&97.60&86.88&80.73\\
RDE~\cite{qin2024noisy} & 87.60&95.65&97.46&86.10&79.54 \\
CFAM(B/16)~\cite{zuo2024ufinebench}&85.55& 94.51 &97.02 &84.23&-\\
CFAM(L/14)~\cite{zuo2024ufinebench}&88.51&95.58 &97.49& 87.09&-\\\hline  
% \rowcolor{green!10} \textcolor{cyan}{\ding{182}}  \textbf{Our ICL}&88.98&96.18&97.76&87.55&81.63\\
% \rowcolor{red!10} \textcolor{cyan}{\ding{182}}  \textbf{Our ICL$\star$}&90.63&96.02&97.76&88.40&82.82\\
% \rowcolor{green!10} \textcolor{cyan}{\ding{183}} \textbf{Our ICL}  &90.81&96.95&98.19&89.58&84.42\\
% \rowcolor{red!10} \textcolor{cyan}{\ding{183}} \textbf{Our ICL$\star$ }&91.70&96.79&98.16&90.21&85.50 \\

\rowcolor{green!10} \textcolor{cyan}{\ding{182}}  \textbf{Our ICL}&89.17&96.13&97.88&87.49&81.50\\
\rowcolor{red!10} \textcolor{cyan}{\ding{182}}  \textbf{Our ICL$\star$}&90.67&95.98&97.86&88.29&82.60 \\
\rowcolor{green!10} \textcolor{cyan}{\ding{183}} \textbf{Our ICL}  &91.02&96.98&98.17&89.76&84.70\\
\rowcolor{red!10} \textcolor{cyan}{\ding{183}} \textbf{Our ICL$\star$ }& 91.78&96.83&98.16&90.33&85.62\\

\hline
\end{tabular}
} 
\caption{Performance comparison on the UFine6926 dataset. The results of IRRA and RDE are reproduced by us.} 
\label{tab_fine}
\vspace{-.6cm}
\end{table}

\begin{table*}[t]
\centering 
\setlength{\abovecaptionskip}{0.1cm}
\resizebox{\textwidth}{!}{
\setlength\tabcolsep{2pt}
\begin{tabular}{l|c|ccccc|ccccc|ccccc}\hline\hline
    \rowcolor{gray!25} &  & \multicolumn{5}{c|}{{CUHK-PEDES}}       & \multicolumn{5}{c|}{{ICFG-PEDES}}       & \multicolumn{5}{c}{{RSTPReid}}                          \\ 
\rowcolor{gray!25}  Methods &Training Sets& Rank-1            & Rank-5           & Rank-10  & mAP   & mINP  & Rank-1   & Rank-5   & Rank-10  & mAP   & mINP & Rank-1         & Rank-5        & Rank-10       & mAP   & mINP  \\\hline 
% CMPM/C   & RN50       & LSTM         & 49.37          & -             & 79.27 & -     & -     & 43.51 & 65.44 & 74.26 & -     & -    & -           & -          & -          & -     & -     \\
\multirow{3}{*}{IRRA~\cite{jiang2023cross}} &CUHK-PEDES&73.38& 89.93 &93.71 &66.13 &50.24&  42.41&62.11&69.62&21.77&1.95&53.25&77.15&85.35&39.63&16.60      \\
&ICFG-PEDES&33.48&56.29&66.33&31.56&19.20&63.46&80.25&85.82&38.06&7.93&45.30&69.25&78.80&36.82&18.38       \\
&RSTPReid&32.80&55.26&65.81&30.29&17.61& 32.30&49.67&57.80&20.54&3.84&  60.20& 81.30 &88.20& 47.17 &25.28      \\\hline

% \multirow{3}{*}{TBPS~\cite{cao2024empirical}} & CUHK-PEDES&73.54& 88.19& 92.35& 65.38& 49.25& 45.53&63.81&71.35&23.73&1.83&53.05&76.15&83.65&38.18&14.90     \\
% &ICFG-PEDES&37.85&58.53&67.74&33.60&20.09&65.05&80.34&85.47&39.83&7.84& 48.35&71.85 &80.20& 38.63& 19.54   \\
% &RSTPReid&28.80& 48.70& 58.67& 26.12& 14.24 &36.34& 53.09 &61.08& 22.75 &4.14&62.10& 81.90& 87.75& 48.00& 25.86     \\\hline

\multirow{3}{*}{RDE~\cite{qin2024noisy}} & CUHK-PEDES&75.94 &90.14& 94.12& 67.56& 51.44&   48.18&66.30&73.70&25.00&2.33&54.90&77.50&86.50&41.27&17.84     \\
&ICFG-PEDES&38.11&59.24&68.44&34.16&20.44&67.68& 82.47& 87.36& 40.06& 7.87 &49.25&72.10&80.20&38.46&18.33  \\
&RSTPReid&  36.94&58.22&67.58&33.65&20.42& 42.17&58.32&65.49&26.37&4.94&65.35 &83.95 &89.90& 50.88 &28.08    \\\hline

% \multirow{3}{*}{\textbf{Our ICL}} & CUHK-PEDES&76.46&90.61 &94.43 &67.96& 51.74&49.33&67.15&74.06&25.72&2.45 &55.30&78.30&87.40&41.71&17.75      \\
% &ICFG-PEDES&45.57&66.46&74.61&40.30&25.46&   67.83& 82.80 &87.66& 41.28& 8.31&55.70&78.55&85.85&44.44&23.09   \\
% &RSTPReid&41.34&62.25&71.00&37.07&22.93&  47.37&63.26&70.09&29.55&5.63&68.25 &86.50 &91.65& 53.04 &29.45 \\\hline
\multirow{3}{*}{\textbf{Our ICL}} & CUHK-PEDES&76.41&90.48&94.33 &68.04&51.99&48.57&66.66&73.75&25.30&2.40& 55.80&79.60&87.65&42.09&17.41  \\
&ICFG-PEDES& 42.87&64.20&73.44&38.19&23.58&68.11&82.59&87.52&40.81&8.18& 52.50&75.05&83.00&41.82&21.14   \\
&RSTPReid&41.31&61.86&70.31&36.78&22.37 &   45.93&62.70&68.80&28.89&5.63&67.70&86.05&91.75&52.62&29.36 \\\hline

\multirow{3}{*}{\textbf{Our ICL$\star$}} & CUHK-PEDES&77.91&90.27&94.14&69.13&53.40 &52.80&66.49&73.49&25.60&2.44& 61.30&79.25&87.40&43.42&18.01       \\
&ICFG-PEDES &49.29&64.34&73.55&40.82&25.38 &69.02&82.45&87.36&41.21&8.30&  60.15&75.30&83.15&43.72&22.04     \\
&RSTPReid& 47.35&61.45&70.34&38.91&23.68 & 50.52&61.56&68.57&29.26&5.73& 70.55&85.95&91.65&53.68&30.13  \\  \hline\hline
\end{tabular}
} 

\caption{Comparison of mutual generalization capabilities between coarse-grained datasets.} 
\label{tab_c_1}
% \vspace{-0.3cm}
\end{table*}

\begin{table}[]
\centering
\setlength{\abovecaptionskip}{0.1cm} 
\resizebox{\linewidth}{!}{
\setlength\tabcolsep{2pt} 
\begin{tabular}{c|l|ccccc}\hline
\rowcolor{gray!25} Source $\rightarrow$ Target & Methods & Rank-1 & Rank-5 & Rank-10 & mAP & mINP \\\hline
\multirow{4}{*}{CUHK.  $\rightarrow$  UFine.} & IRRA~\cite{jiang2023cross}&  37.51 & 54.92 & 64.29 & 40.76 & 34.33 \\
 & RDE~\cite{qin2024noisy} & 40.37 & 57.49 & 66.05 & 42.68 & 35.78 \\
 &\textbf{Our ICL} & 46.40 & 63.55 & 72.08 & 48.68 & 41.56 \\
 & \textbf{Our ICL$\star$} & \textbf{57.76}&\textbf{64.13}&\textbf{72.81}&\textbf{53.97}&\textbf{45.64} \\\hline
\multirow{4}{*}{ICFG.  $\rightarrow$  UFine.} & IRRA~\cite{jiang2023cross}&  15.02 & 26.79 & 33.90 & 17.10 & 12.75 \\
 & RDE~\cite{qin2024noisy} & 17.86 & 31.01 & 38.56 & 19.82 & 14.74 \\
 &\textbf{Our ICL} & 27.95 & 44.20 & 52.20 & 29.85 & 23.20 \\
 & \textbf{Our ICL$\star$} &  \textbf{36.81}&\textbf{44.65}&\textbf{52.73}&\textbf{34.12}&\textbf{26.61}  \\\hline
\multirow{4}{*}{RSTP.  $\rightarrow$  UFine.} & IRRA~\cite{jiang2023cross}&  13.21 & 25.67 & 33.93 & 15.60 & 11.09 \\
 & RDE~\cite{qin2024noisy} & 14.00 & 25.23 & 32.64 & 16.22 & 11.90 \\
 &\textbf{Our ICL} & 23.89 & 38.30 & 46.70 & 25.54 & 19.20 \\
 & \textbf{Our ICL$\star$} & \textbf{31.23}&\textbf{38.56}&\textbf{47.02}&\textbf{28.90}&\textbf{21.80} \\\hline\hline

\multirow{4}{*}{UFine.  $\rightarrow$  CUHK} & IRRA~\cite{jiang2023cross}&  37.74 & 60.12 & 70.13 & 35.94 & 23.21 \\
 & RDE~\cite{qin2024noisy} & 39.41 & 61.14 & 70.11 & 36.49 & 23.32 \\
 &\textbf{Our ICL} & 49.04 & 70.27 & \textbf{78.64} & 44.54 & 29.58 \\
 & \textbf{Our ICL$\star$} &  \textbf{56.87}&\textbf{70.19}&{78.53}&\textbf{47.31}&\textbf{31.20}\\\hline
\multirow{4}{*}{UFine.  $\rightarrow$  ICFG.} & IRRA~\cite{jiang2023cross}&  34.52 & 55.41 & 64.44 & 17.96 & 1.95 \\
 & RDE~\cite{qin2024noisy} & 40.37 & 60.14 & 68.41 & 20.54 & 2.19 \\
 &\textbf{Our ICL} & 43.10 & \textbf{62.92} & \textbf{70.73} & 22.73 & 2.56 \\
 & \textbf{Our ICL$\star$} &  \textbf{47.83}&62.67&70.48&\textbf{23.16}&\textbf{2.62}  \\\hline
\multirow{4}{*}{UFine.  $\rightarrow$  RSTP.} & IRRA~\cite{jiang2023cross}&  37.65 & 63.70 & 73.00 & 29.00 & 11.80 \\
 & RDE~\cite{qin2024noisy} & 39.90 & 63.50 & 74.75 & 29.92 & 12.43 \\
 &\textbf{Our ICL} & 48.85& \textbf{72.65} & \textbf{81.80} & 36.91 & 16.39 \\
 & \textbf{Our ICL$\star$} &\textbf{55.35}&72.40&81.50&\textbf{38.64}&\textbf{17.23} \\\hline
\end{tabular}}
\caption{Generalization capabilities between coarse-grained and fine-grained datasets. The best scores in each task are in \textbf{bold}.} 
\label{tab_c_2}
% \vspace{-0.3cm}
\end{table}

\subsection{Comparison with State-of-the-Arts \label{sec4.1}}
In this section, to verify the superiority of our ICL, we compare our method with more than 15 baselines including recent advanced methods (\eg, RDE (CVPR'24)~\cite{qin2024noisy} and NAM (CVPR'24)~\cite{tan2024harnessing}). Based on the backbone type, we divide the baselines into two groups (\textcolor{cyan}{\ding{182}} and \textcolor{cyan}{\ding{183}}) as shown~\Cref{tab_main}, \ie, the baselines using VL-Backbones w/o and with ReID-domain pre-training. For the group \textcolor{cyan}{\ding{183}}, we use the pre-trained weights released by~\cite{tan2024harnessing} to initialize CLIP for fair comparison. The results are reported in~\Cref{tab_main,tab_fine}.

\noindent\textbf{Results on coarse-grained datasets.} \Cref{tab_main} report the results evaluated on the coarse-grained datasets. We can see that our method can achieve competitive performance even without THI. By performing THI, the Rank-1 score of our method is greatly improved, \eg, in the group \textcolor{cyan}{\ding{182}}, the Rank-1 scores on the three datasets are improved by 1.50\%, 0.91\%, and 2.85\%, respectively. In addition, mAP and mINP scores have also improved greatly, which indicates that the overall ranking has improved.
In group \textcolor{cyan}{\ding{183}}, our method achieves the best scores on most metrics, especially Rank-1 reached 72.55\% on RSTPReid, which is sufficient to verify the superiority. However, we found that THI slightly degraded Rank-5 and Rank-10, which is because we only conducted 5 rounds of interaction. As long as the locating anchor image is incorrect, the errors will accumulate, which can be relieved by increasing the rounds of interaction. But in general, THI can significantly improve Rank-1 and the overall ranking (mAP and mINP).

\noindent\textbf{Results on fine-grained dataset.}  \Cref{tab_fine} shows the results on the fine-grained dataset whose average text length is over $80$. Such kind of fine-grained description often has a clear query intention. For better comparison, we provide the performance of IRRA and RDE on the UFine6926 dataset. From the results, our method can still achieve excellent performance, with Rank-1 exceeding 91\%. This shows that interaction is also applicable to the fine-grained scenario.

\subsection{Generalization Study \label{sec4.2}}
To evaluate the generalization, \Cref{tab_c_1,tab_c_2} report the cross-domain performance on four datasets, including the coarse-to-fine, coarse-to-fine, and fine-to-coarse generalization experiments. From the results, our method achieves better generalization than RDE even without THI thanks to the training enhancement of RDA. When THI is performed, the cross-domain performance is dramatically improved, for example, from CUHK-PEDES to RSTPReid, THI brings an improvement of more than 4\% on Rank-1. This shows that THI is a potential solution to generalization challenges in the future. From the generalization experiments between fine-grained and coarse-grained datasets shown in~\Cref{tab_c_2}, ICL can also achieve the best cross-domain performance, \eg, compared with the best baseline RDE, from UFine6926 domain to CUHK-PEDES domain, our method improves Rank-1 and mAP by 17.49\% and 10.86\%, respectively, which further verifies the cross-domain generalization of our method. 

% In addition, some recent studies have shown that pre-training can effectively improve cross-domain generalization capabilities. To verify the effectiveness and advantages of our method, we use the ReID-domain pre-training weights published by NAM~\cite{tan2024harnessing} to initialize the model and conduct the cross-domain evaluation of three downstream coarse-grained datasets. The experimental results are as presented in~\Cref{tab_pre}.

\begin{table}[!hbtp]
\centering
\setlength{\abovecaptionskip}{0.1cm}
\arrayrulecolor{black}
\resizebox{\linewidth}{!}{
\setlength\tabcolsep{3pt}
\begin{tabular}{l|c|cc|cc|cc|c}\hline
\rowcolor{gray!25}    & & \multicolumn{2}{c|}{{CUHK-PEDES}} & \multicolumn{2}{c|}{{ICFG-PEDES}}  & \multicolumn{2}{c|}{{RSTPReid}} & \\ 
 \rowcolor{gray!25} Methods&THI& Rank-1 & mAP &Rank-1 & mAP &Rank-1 & mAP& $\Delta$Avg \\\hline  
\multirow{2}{*}{CLIP~\cite{radford2021learning}} &\textcolor{red}{\ding{56}}&71.64&63.92&60.11&34.52&56.55&44.52&\multirow{2}{*}{\textbf{+2.57}} \\
&\textcolor{green}{\ding{52}}&73.77&65.66&63.57&34.95&61.45&46.25   \\\hline

\multirow{2}{*}{IRRA~\cite{jiang2023cross}} &\textcolor{red}{\ding{56}}&73.38&66.13&63.46&38.06&60.20&47.17&\multirow{2}{*}{\textbf{+1.86}}\\
&\textcolor{green}{\ding{52}}&76.06&67.42&65.26&38.58 &63.75&48.47   \\\hline

\multirow{2}{*}{RDE~\cite{qin2024noisy}} &\textcolor{red}{\ding{56}}&75.94& 67.56&67.68&40.06&65.35&50.88&\multirow{2}{*}{\textbf{+1.41}}\\
&\textcolor{green}{\ding{52}}& 77.47&68.62&68.72&40.63&68.45&52.01  \\
\hline
\end{tabular}
}
\caption{Transferability results on three coarse-grained benchmarks. $\Delta$Avg represents the average improvement.} 
\label{tab_trans}
% \vspace{-0.3cm}
\end{table}

\subsection{Transferability Study \label{sec4.4}}
Our interactive module is separate and independent from the training of existing TIReID methods, so it can be plug-and-play with existing methods to improve ReID performance. To verify the transferability of our THI, we conduct experiments on multiple baselines, and the results are shown in~\Cref{tab_trans}. Except for the CLIP results we reproduced, all other experiments used public pre-trained model weights. From the results of \Cref{tab_trans}, the interactive strategy application can significantly improve Rank-1 and mAP, which shows that the external guidance by interactions via MLLMs can further clarify the text-image alignments and improve the overall ranking. The average improvements on the three datasets are 2.57\%, 1.86\%, and 1.41\%, respectively, which seriously proves the transferability of our THI.

\subsection{Ablation and Parameter Study \label{sec4.3}}
To explore the effects of each proposed component, \ie, THI and RDA, we first conduct the ablation study on three coarse-grained datasets as reported in~\Cref{tab_ab}. From the results, each component can bring performance gains on Rank-1 and mAP, which verifies the reliability and rationality of the method. Especially the introduction of THI has greatly improved the Rank-1 accuracy from 68.95\% to 70.55\% on the RSTPReid dataset. In addition, our RDA also brings considerable performance gains, especially on the RSTPreid dataset. Also, we conduct the parameter analysis on the CHUK-PEDES dataset on two free hyperparameters, \ie, the similarity threshold $\xi$ and the balance factor $\lambda$. The former filters unnecessary retrievals to improve the interaction efficiency, while the latter controls the contribution of refined texts. Based on~\Cref{fig_para}, in our experiments, we set $\xi$ in the range of $0.5\sim0.6$ and $\lambda$ to $0.8$, thus mitigating the risk of introducing noisy external knowledge.

% Based on~\Cref{fig_para},  in our experiments, we set $\xi$ within $0.45\sim0.55$ and  $\lambda$ to $0.8$  for the mitigation of risks introducing noisy external knowledge, respectively.

% when $0.7<\xi<0.8$, the optimal performance is obtained. Our method achieves the best performance when $\lambda=0.8$, because a too-small value will cause query drift,  causing suboptimal performance. In all experiments, $\xi$ and $\lambda$ are set to $0.5$ and $0.8$, respectively.

% we can see that: \textcolor{green}{(1)} Placeholder Placeholder Placeholder Placeholder Placeholder Placeholder Placeholder Placeholder Placeholder Placeholder Placeholder Placeholder Placeholder Placeholder Placeholder Placeholder Placeholder Placeholder Placeholder Placeholder Placeholder Placeholder Placeholder Placeholder Placeholder Placeholder Placeholder Placeholder Placeholder Placeholder Placeholder Placeholder Placeholder Placeholder Placeholder Placeholder Placeholder Placeholder Placeholder Placeholder Placeholder Placeholder Placeholder.

\begin{table}[!hbtp]
\centering
\setlength{\abovecaptionskip}{0.1cm}
\resizebox{\linewidth}{!}{
\setlength\tabcolsep{2pt}
\begin{tabular}{c|ccc|cc|cc|cc}\hline
\rowcolor{gray!25}  & && & \multicolumn{2}{c|}{{CUHK-PEDES}} & \multicolumn{2}{c|}{{ICFG-PEDES}}  & \multicolumn{2}{c}{{RSTPReid}}  \\ 
 \rowcolor{gray!25} No.&THI&RDA&LoRA& Rank-1 & mAP &Rank-1 & mAP &Rank-1 & mAP\\\hline  
\#1&\textcolor{green}{\ding{52}}& \textcolor{green}{\ding{52}}&\textcolor{green}{\ding{52}}& \textbf{ 77.91}&\textbf{69.32}&\textbf{69.02}&\textbf{41.21}&\textbf{70.55}&\textbf{53.68}  \\
\#2&\textcolor{green}{\ding{52}}& \textcolor{green}{\ding{52}}&\textcolor{red}{\ding{56}}& 76.38&68.59& 67.92  & 41.13&69.00 &53.11   \\
% \#3&\textcolor{green}{\ding{52}}& \textcolor{red}{\ding{56}} &\textcolor{green}{\ding{52}}&  \\
\#3&\textcolor{red}{\ding{56}} & \textcolor{green}{\ding{52}}&\textcolor{red}{\ding{56}}&76.41&68.04&68.11&40.81&67.70&52.62  \\
% \#5&\textcolor{red}{\ding{56}} &\textcolor{red}{\ding{56}} &\textcolor{red}{\ding{56}}& 77.73&69.47&50.94&26.79&57.90&43.46 \\
\#4&\textcolor{red}{\ding{56}} &\textcolor{red}{\ding{56}} &\textcolor{red}{\ding{56}}& 75.94&67.56&67.68&40.06&65.35&50.88 \\
\hline
\end{tabular}
}
% \caption{Ablation studies the CHUK-PEDES dataset. We train the model on the CHUK-PEDES dataset and separately evaluate the in- and cross-domain performance.} 
\caption{Ablation studies on CHUK-PEDES, ICFG-PEDES, and RSTPReid datasets. The best scores are in \textbf{bold}.} 
\label{tab_ab}
\vspace{-0.3cm}
\end{table}

\begin{figure}[!hbtp]
\centering
\setlength{\abovecaptionskip}{0.1cm}
\begin{subfigure}{0.495\linewidth}
\hfill
\includegraphics[width=\linewidth]{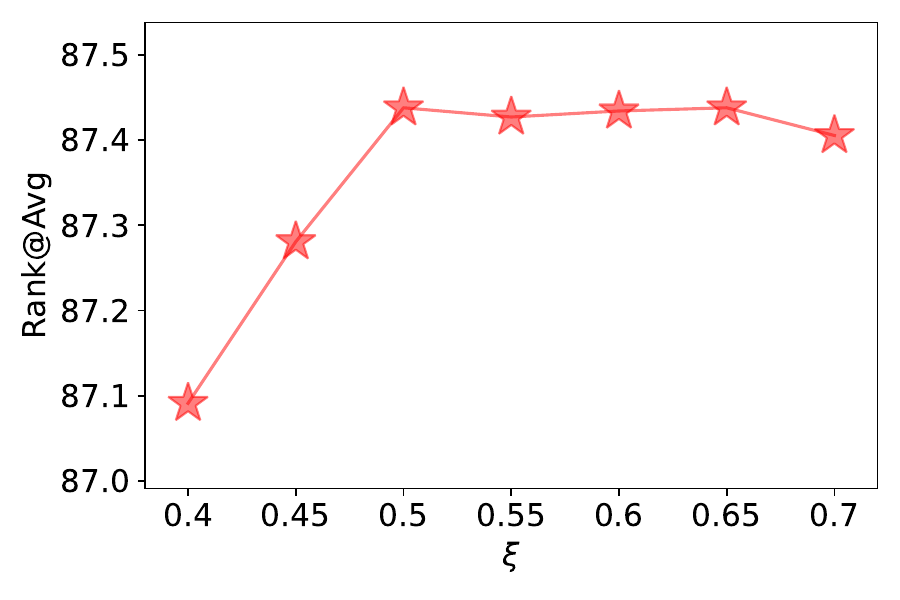}
\centering
\caption{The similarity 
threshold $\xi$}
\label{cuhk_test_sims}
\end{subfigure}
\begin{subfigure}{0.495\linewidth}
\hfill
\includegraphics[width=\linewidth]{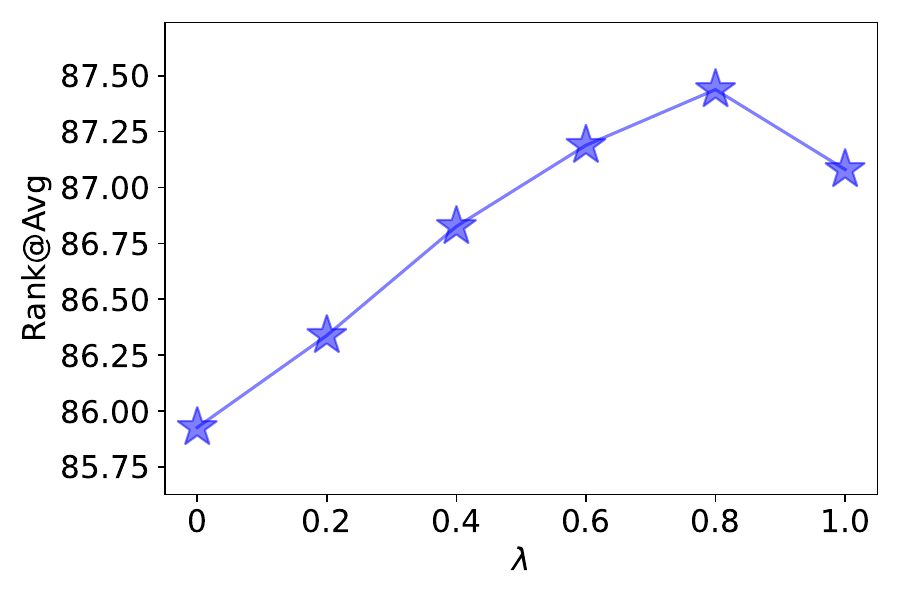}
\centering
\caption{The balance factor $\lambda$}
\label{icfg_test_sims}
\end{subfigure}
% \begin{subfigure}{0.32\linewidth}
% \hfill
% \includegraphics[width=\linewidth]{img/sims_test.pdf}
% \centering
% \caption{\footnotesize  {RSTPReid}}
% \label{rstp_test_sims}
% \end{subfigure}
\caption{ Variation of performance with different $\xi$ and $\lambda$.}
\label{fig_para} 
\vspace{-0.3cm}
\end{figure}

\subsection{Interactive Study \label{sec4.5}}
\begin{figure}[t]
\centering
\setlength{\abovecaptionskip}{0.1cm}
\begin{subfigure}{0.325\linewidth}
\includegraphics[width=\linewidth]{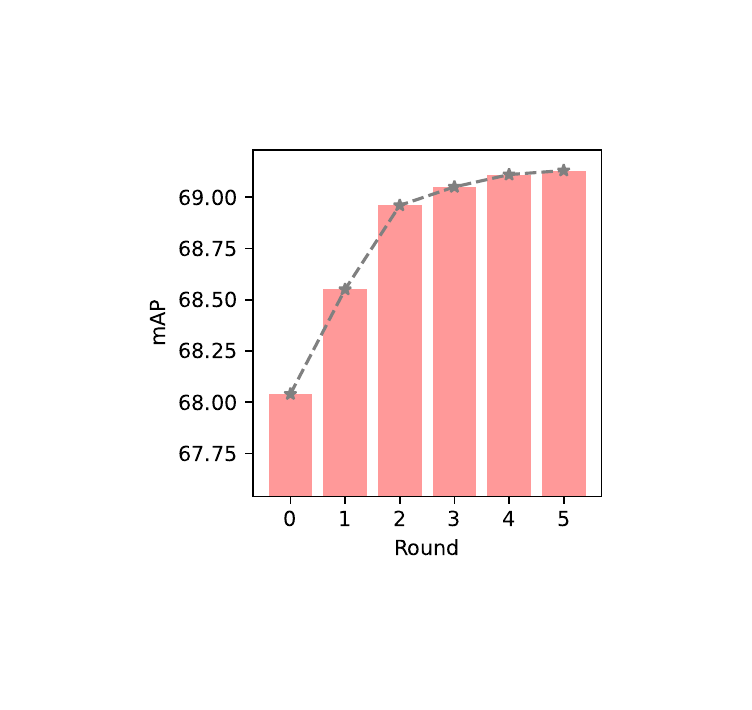}
\centering
\caption{CUHK-PEDES}
\label{cuhk_test_sims}
\end{subfigure}
\begin{subfigure}{0.325\linewidth}
\includegraphics[width=\linewidth]{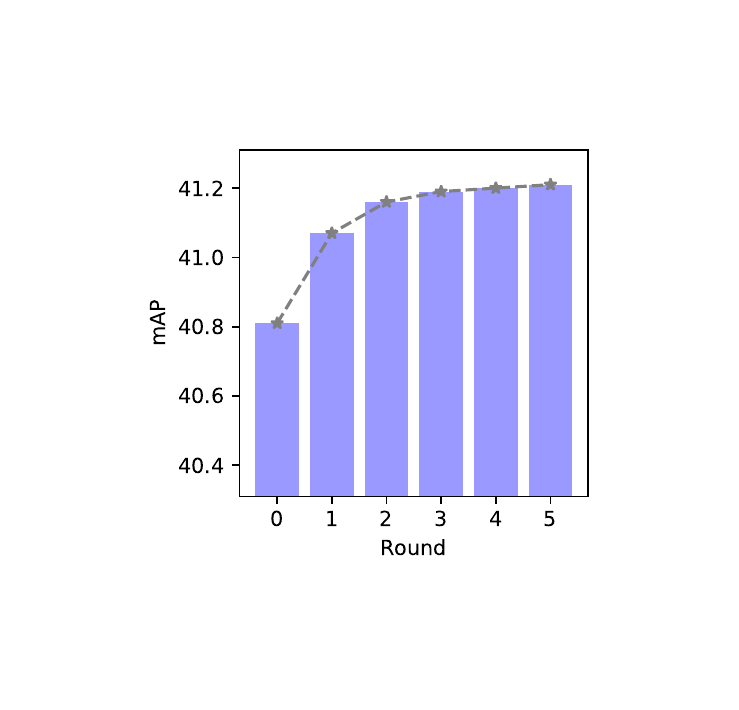}
\centering
\caption{ICFG-PEDES}
\label{icfg_test_sims}
\end{subfigure} 
\begin{subfigure}{0.325\linewidth}
\includegraphics[width=\linewidth]{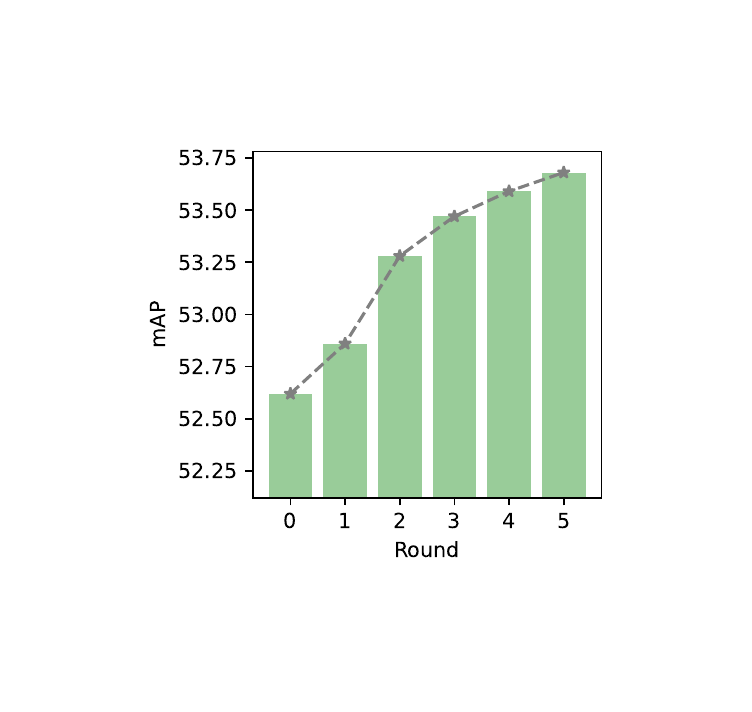}
\centering
\caption{RSTPReid}
\label{icfg_test_sims}
\end{subfigure} 
\caption{Performance (mAP) versus rounds on three datasets. Round 0 indicates the setting without using THI.}
\label{fig_round}  
\end{figure}

\begin{figure}[t]
\centering
\setlength{\abovecaptionskip}{0.1cm}
\resizebox{\linewidth}{!}{ 
\includegraphics{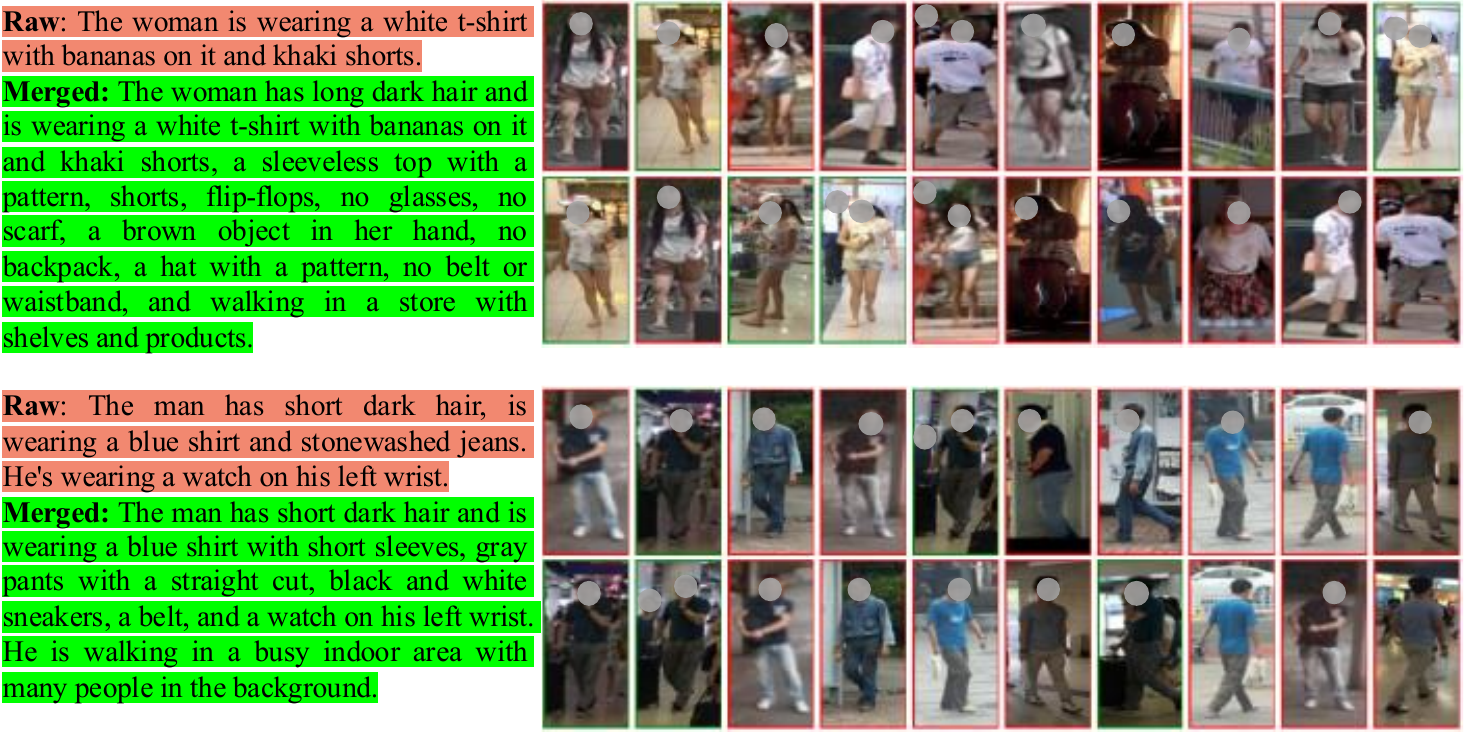}}
\caption{Top-10 retrieved results on CUHK-PEDES dataset between ICL (the first row) and ICL with THI (the second row).}
\label{fig_egg} 
\end{figure}

This section further explores the interactive module, \ie THI. We report the performance (mAP) changes after multiple rounds of interactions in~\Cref{fig_round}.  The mAP score can reflect the overall retrieval quality.  From the results, the performance improvement is obvious in the first few rounds ($\leq 2$) since the items that meet the query semantics are mostly concentrated at the top of the ranking.  However, after $>2$ rounds, the performance gain is gradually not obvious as the number of queries requiring interaction decreases.  But generally, as the number of rounds increases, the overall performance gradually improves. In all our experiments, we performed 5 rounds of interaction. In addition, we visualize the top-10 retrieved results before and after applying THI in~\Cref{fig_egg}. Due to the inherent limitations of the intra-model knowledge, it is difficult to obtain satisfactory results without the help of THI. In contrast, our THI can dynamically enrich queries by interacting with MLLMs to achieve a more reliable ranking. More example results are given in the supplementary material.

\section{Conclusion}
In this paper, we explore interactive text-to-image person re-identification, which aims to improve the alignment between dynamic queries and challenging candidate images by leveraging external guidance from MLLMs. To achieve this, we develop an Interactive Cross-modal Learning (ICL) framework to alleviate the inherent challenges of offline models and training data by, including a plug-and-play Test-time Human-centered Interaction (THI) module and Reorganization Data Augmentation (RDA). Extensive experiments and analysis show that our framework can effectively transfer external knowledge in MLLMs into offline models for guiding re-identification, showing excellent performance and generalization.

\section*{Acknowledgments}
This work was supported in part by the National Key R\&D Program of China under Grant 2024YFB4710604; in part by NSFC under Grants 62472295, 62372315, 62176171, and U21B2040; in part by Sichuan Science and Technology Planning Projects (2024YFHZ0089, 2024NSFTD0049, 2024YFHZ0144, 2024NSFTD0047, 2024NSFTD0038); in part by System of Systems and Artificial Intelligence Laboratory pioneer fund grant; in part by the Fundamental Research Funds for the Central Universities under Grant CJ202403 and CJ202303; in part by Chengdu Science and Technology Project under Grant 2023-XT00-00004-GX.

% This work was supported in part by NSFC under Grant U21B2040, 62176171, 62372315, and 62102274, in part by Sichuan Science and Technology Program under Grant 2022YFH0021 and 2023ZYD0143; in part by Chengdu Science and Technology Project under Grant 2023-XT00-00004-GX; in part by the SCU-LuZhou Sciences and Technology Coorperation Program under Grant 2023CDLZ-16; in part by the Fundamental Research Funds for the Central Universities under Grant CJ202303 and YJ202140.
{
    \small
    \bibliographystyle{ieeenat_fullname}
    \bibliography{main}
}

% WARNING: do not forget to delete the supplementary pages from your submission 
% \input{sec/X_suppl}

\end{document}